\documentclass{article}

\PassOptionsToPackage{sort&compress}{natbib}

\usepackage[preprint]{neurips_2024}

\usepackage[utf8]{inputenc} 
\usepackage[T1]{fontenc}    
\usepackage[colorlinks=true, allcolors=blue]{hyperref}       
\usepackage{url}            
\usepackage{booktabs}       
\usepackage{amsfonts}       
\usepackage{nicefrac}       
\usepackage{microtype}      
\usepackage[dvipsnames]{xcolor}         

\usepackage{graphicx}
\usepackage{subcaption}
\usepackage{lipsum}
\usepackage{float}

\usepackage{booktabs}
\usepackage{xfrac}
\usepackage{amsmath}
\usepackage{multicol}
\usepackage{wrapfig}

\usepackage[capitalise,noabbrev]{cleveref}
\creflabelformat{equation}{#2\textup{#1}#3}

\usepackage{listings}
\definecolor{backcolor}{rgb}{0.95,0.95,0.92}
\usepackage{circledsteps}
\pgfkeys{/csteps/inner color=white}
\pgfkeys{/csteps/fill color=orange}
\pgfkeys{/csteps/outer color=orange}

\newif\ifdraft 
\drafttrue

\title{Four Over Six: More Accurate NVFP4 Quantization with Adaptive Block Scaling}

\author{%
  Jack Cook$^1$\thanks{Work was partly done during an internship at NVIDIA. $^1$Massachusetts Institute of Technology $^2$NVIDIA.} \\
  \And
  Junxian Guo$^1$ \\
  \And
  Guangxuan Xiao$^1$ \\
  \And
  Yujun Lin$^2$ \\
  \AND
  Keith Wyss$^2$ \\
  \And
  Mahdi Nazemi$^2$ \\
  \And
  Asit Mishra$^2$ \\
  \And
  Carlo del Mundo$^2$ \\
  \And
  Tijmen Blankevoort$^2$ \\
  \And
  Song Han$^{1,2}$
}

\begin{document}

\maketitle

\begin{abstract}
As large language models have grown larger, interest has grown in low-precision numerical formats such as NVFP4 as a way to improve speed and reduce memory usage.
However, quantizing models to NVFP4 remains challenging as the lack of precision generally degrades model performance.
In this work, we address this issue with \textit{Four Over Six (4/6)}, a modification to the block-scaled NVFP4 quantization algorithm that yields reduced quantization error.
Unlike integer formats, floating point formats have non-uniform step sizes which create larger quantization error on larger values.
4/6 takes advantage of this by adaptively scaling some blocks to smaller FP4 values, making the distribution of representable values more uniform and reducing quantization error for near-maximal values.
We show that 4/6 can be implemented efficiently on modern hardware accelerators, resulting in performance gains during both pre-training and inference with minimal computational overhead.
In pre-training experiments with the Nemotron 3 Nano 30B-A3B model architecture, we find that 4/6 brings training loss closer to BF16 compared to models trained with current state-of-the-art NVFP4 training recipes.
\end{abstract}

\section{Introduction}

Large language models (LLMs) have grown increasingly capable as a direct result of increases to their sizes and the speeds at which they can be trained.
As a result, creating methods that enable the training of larger LLMs has been a driving question behind machine learning systems research for the past several years.
An example of this can be seen in numerical precision: it used to be standard to train models using FP32, then FP16~\cite{micikevicius_mixed_2018}, and now it is standard to keep most or all parameters in BF16~\cite{kalamkar_study_2019}.
Some works have successfully trained LLMs while keeping some parameters in FP8, but this remains an active area of study~\cite{deepseek-ai_deepseek-v3_2025,meta_ai_llama_2025,mishra_recipes_2025}.

Going beyond FP8, some works have begun to study the feasibility of training LLMs with FP4~\cite{castro_quartet_2025,chmiel_fp4_2025,tseng_training_2025,nvidia_pretraining_2025}.
However, end-to-end training with FP4 remains challenging, as FP4 is a very coarse datatype that only has 16 values: $\pm\{0, 0.5, 1, 1.5, 2, 3, 4, 6\}$.
To make up for this lack of precision, block-scaled FP4 formats such as MXFP4~\cite{rouhani_microscaling_2023} and NVFP4~\cite{nvidia_pretraining_2025} have recently become more popular.
Rather than simply quantizing all values in a tensor to the range of FP4 from $-6$ to $6$, block-scaled formats allow one FP8 scale factor to be stored for every $m$ values, 32 and 16 for MXFP4 and NVFP4 respectively, enabling a much larger range of values to be represented across an entire tensor.

Even with this change, using NVFP4 for training remains challenging: current hardware accelerators that support NVFP4, such as NVIDIA Blackwell GPUs, require that both operands of any matrix multiplication are quantized to NVFP4.
As a result, for efficient model training, weights, activations, and gradients must all be quantized to NVFP4.
In theory, this should deliver two key benefits: speed improvements due to the 4-6x faster matrix multiplication operations compared to BF16~\cite{nvidia_pretraining_2025}, and a natively quantized NVFP4 model.
In practice, however, current state-of-the-art NVFP4 training recipes require operations that introduce more computational overhead and fail to deliver a natively quantized model.
These include the random Hadamard transform (RHT), stochastic rounding (SR), keeping some layers in high precision, and ``healing'' the model by switching to high precision weights, activations, and gradients near the end of training~\cite{chmiel_fp4_2025,nvidia_pretraining_2025,wang_optimizing_2025,castro_quartet_2025}.
Furthermore, this added overhead must be carefully managed: if too much overhead is introduced, it becomes faster to train models using more accurate FP8 formats.
To make NVFP4 training viable, more lightweight operations that improve numerical accuracy are necessary.

\begin{figure*}
    \centering
    \begin{subfigure}[t]{2.65in}
        \centering
        \includegraphics[width=\linewidth]{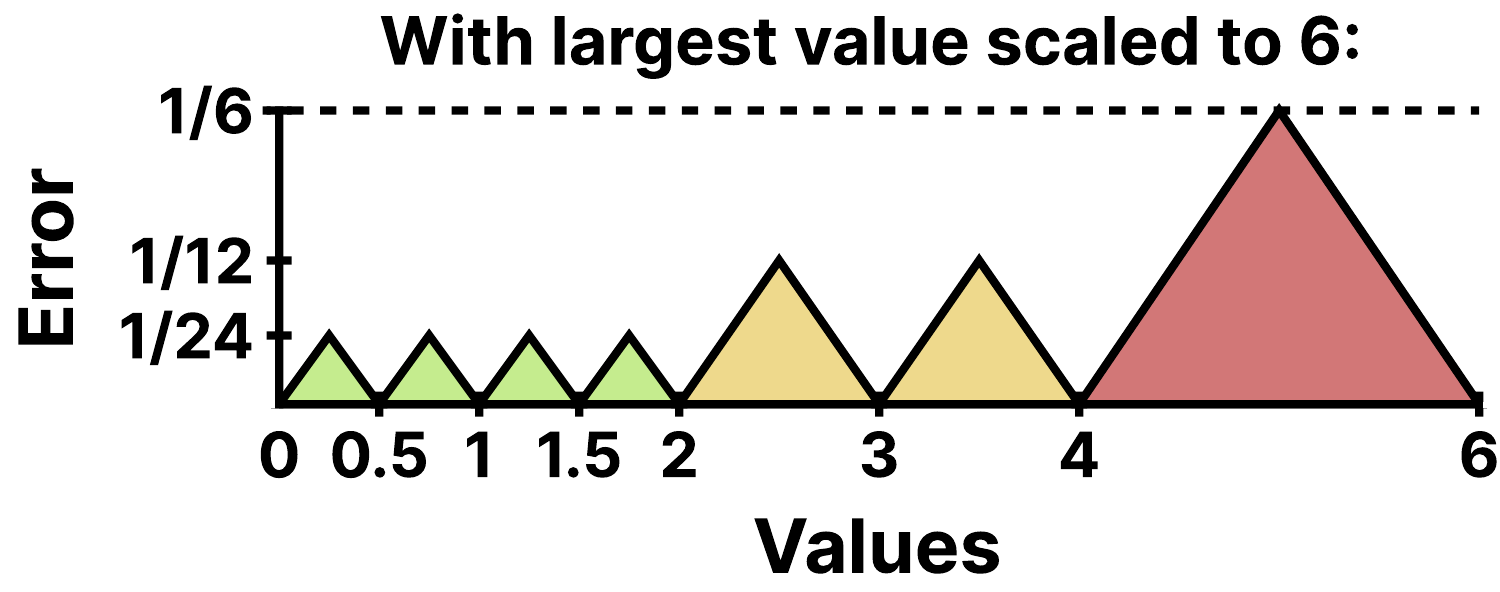}
        \caption{Quantization error relative to the largest value in a block of values quantized to NVFP4 when the largest value in the block is scaled to six.}
        \label{fig:scaled-to-6}
    \end{subfigure}
    \hfill
    \begin{subfigure}[t]{2.65in}
        \centering
        \includegraphics[width=\linewidth]{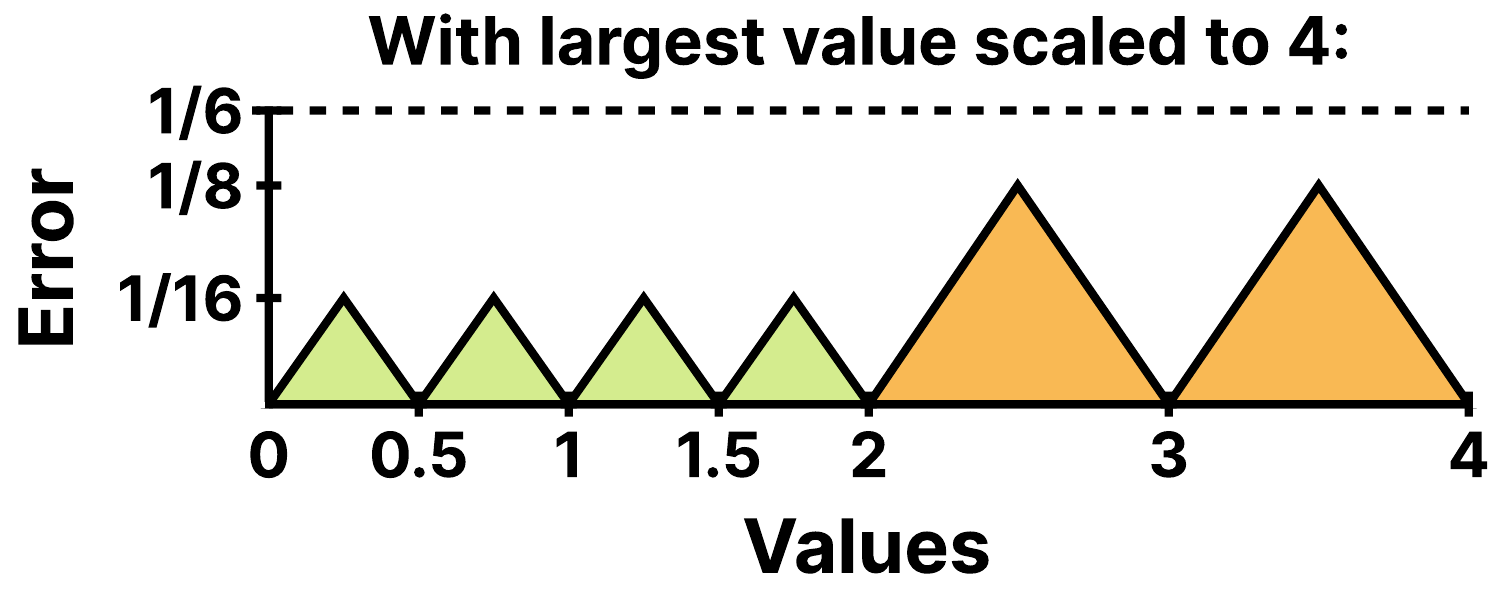}
        \caption{Quantization error relative to the largest value in the block when the block's largest value is instead scaled to four.}
        \label{fig:scaled-to-4}
    \end{subfigure}
    \caption{In standard NVFP4 quantization (left), using the full range of FP4 values from 0 to 6 means that it is impossible to represent values between 66.6\% and 100\% of the magnitude of the largest value in a block. By instead scaling some blocks to a maximum value of 4, it becomes possible to represent values that are 75\% of the largest value in a block, reducing worst-case quantization error for large values.}
    \label{fig:4-over-6}
\end{figure*}

\begin{wraptable}{r}{2.65in}
    \centering
    \begin{tabular}[b]{lr}
        \toprule
                           & [10, 20, 30, 40]                                    \\
        \midrule
        NVFP4 ($M=6$)        & \Circled{6.5} $\times$ {[}1.5, 3, \Circled{4}, 6{]} \\
        Mean Squared Error & 4.33                                                \\
        \midrule
        NVFP4 ($M=4$)        & \textbf{10 $\times$ {[}1, 2, 3, 4{]}}               \\
        Mean Squared Error & \textbf{0}                                          \\
        \bottomrule
    \end{tabular}
    \caption{NVFP4 quantization using \cref{eqn:fp4-quantization} with the block's largest value scaled to $M=4$ and $M=6$. Blocks such as this one, often those with values close to 75\% of the block's largest value, may be represented with less error when scaled to 4. See \cref{tab:sample-quantization} for details.}
    \label{tab:nvfp4-4-vs-6-example-small}
\end{wraptable}

In this work, we introduce \textit{Four Over Six (4/6)}, a change to the NVFP4 block-scaled quantization algorithm that improves its representation of near-maximal values in each block.
During quantization, it is standard to scale high-precision values to the full range of values that can be represented by the low-precision numerical format~\cite{nagel_white_2021}.
However, when this is done with FP4, this results in large amounts of quantization error for near-maximal values (\cref{fig:scaled-to-6}).
If values are instead scaled to the range $(-4, 4)$, the distribution of representable values becomes more uniform, reducing worst-case quantization error (\cref{fig:scaled-to-4}).
An example can be seen in \cref{tab:nvfp4-4-vs-6-example-small}.
When the values [10, 20, 30, 40] are quantized to NVFP4, 30 is scaled to $\frac{30}{6.5}=4.62$, and then rounded down to 4, the nearest FP4 value, introducing a relative error of 13.4\%.
If this block is instead scaled such that the largest value is 4, 30 would instead be scaled to $\frac{30}{10}=3$, which can be represented in FP4 with no error.

We find that for many blocks in weight, gradient, and activation tensors during both pre-training and post-training quantization (PTQ), scaling to 4 rather than 6 introduces less error, leading to more accurate models.
Crucially, we find that this can be implemented efficiently in an online fashion, adding less than 15\% overhead to the NVFP4 quantization operation.
During pre-training, we find that 4/6 improves the performance of current state-of-the-art NVFP4 pre-training recipes, bringing loss closer to high-precision baselines.
Furthermore, when used during post-training quantization, we find that 4/6 can often improve the accuracy of existing methods such as GPTQ~\cite{frantar_gptq_2023}, AWQ~\cite{lin_awq_2024}, and SmoothQuant~\cite{xiao_smoothquant_2024}.
We additionally release quantization and matrix multiplication kernels on \href{https://github.com/mit-han-lab/fouroversix}{GitHub}.
\section{Challenges with NVFP4 Quantization}

\subsection{NVFP4 Overview}

NVFP4 is a block-scaled quantization format that stores values in FP4 E2M1 with an FP8 E4M3 scale factor $\Delta_i$ for every 16 values and a tensor-wide FP32 scale factor $\alpha$.
This can be expressed as follows, where $\mathbf{X}$ is the high-precision tensor, $\mathbf{\bar{X}}$ is its quantized representation, $M^\text{FP4}$ and $M^\text{FP8}$ are the largest values that can be represented in FP4 and FP8 E4M3, 6 and 448 respectively, and $\lceil \cdot \rfloor$ is the rounding function.\footnote{FP32 and FP8 casting operations are not described mathematically for brevity. Sample calculations can be found in lines 1-4 of \cref{tab:sample-quantization}.}

\newcommand{\hp}{\mathbf{X}}
\begin{align}
    \alpha^{\text{FP32}} &= \frac{\text{max}(|\hp|)}{M^\text{FP4} \times M^\text{FP8}} \label{eqn:fp4-quantization-tensor-scale} \\
    \Delta^{\text{FP8}}_i           & = \frac{\text{max}(|\hp_{16i...16(i+1)}|)}{\alpha M^\text{FP4}} \label{eqn:fp4-quantization-block-scale} \\
    \mathbf{\bar{X}}^{\text{FP4}} & =
    \begin{cases}
        \frac{1}{2}\lceil \frac{2\hp}{\alpha\Delta} \rfloor, & |\frac{\hp}{\alpha\Delta}| < 2    \\
        \lceil \frac{\hp}{\alpha\Delta} \rfloor,             & |\frac{\hp}{\alpha\Delta}| < 4    \\
        2\lceil\frac{\hp}{2\alpha\Delta}\rfloor,             & |\frac{\hp}{\alpha\Delta}| \leq 6
    \end{cases}
    \label{eqn:fp4-quantization}
\end{align}

Unlike integer formats, floating point formats such as FP4 have dynamic step sizes: 0.5 between 0 and 2, 1 between 2 and 4, and 2 between 4 and 6, as shown in \cref{fig:scaled-to-6}.
This allows them to represent a wider range of values, making them better at representing weights, activations, and gradients in many cases~\cite{chen_int_2025,liu_llm-fp4_2023}.
For example, the range of representable values, which is the largest value divided by the smallest positive value, for FP4 is 12 ($\frac{6}{0.5}$), compared to just 7 ($\frac{7}{1}$) for INT4.

Note that \cref{eqn:fp4-quantization-block-scale,eqn:fp4-quantization} involve casting values to specific numerical precisions.
These are required to benefit from hardware support, but they are also where quantization error is introduced, as information is lost when rounding to the nearest representable values.
In summary, there are two sources of error in NVFP4 quantization:

\begin{enumerate}
    \item \textbf{FP8 block-level scale factors:} Each scale factor contains some quantization error due to the limited precision offered by E4M3. If a scale factor is rounded up or down from its high-precision counterpart, it affects all of the values in its block.
    \item \textbf{FP4 values:} Individual values may be rounded up or down to one of the eight possible FP4 values shown in \cref{fig:scaled-to-6}.
\end{enumerate}

\subsection{NVFP4 Error Comes From Rounding Near-Maximal Values}
\label{sec:quantizing-llms-to-nvfp4}

\begin{figure}
    \centering
    \begin{subfigure}[t]{0.44\textwidth}
        \centering
        \includegraphics[width=\linewidth]{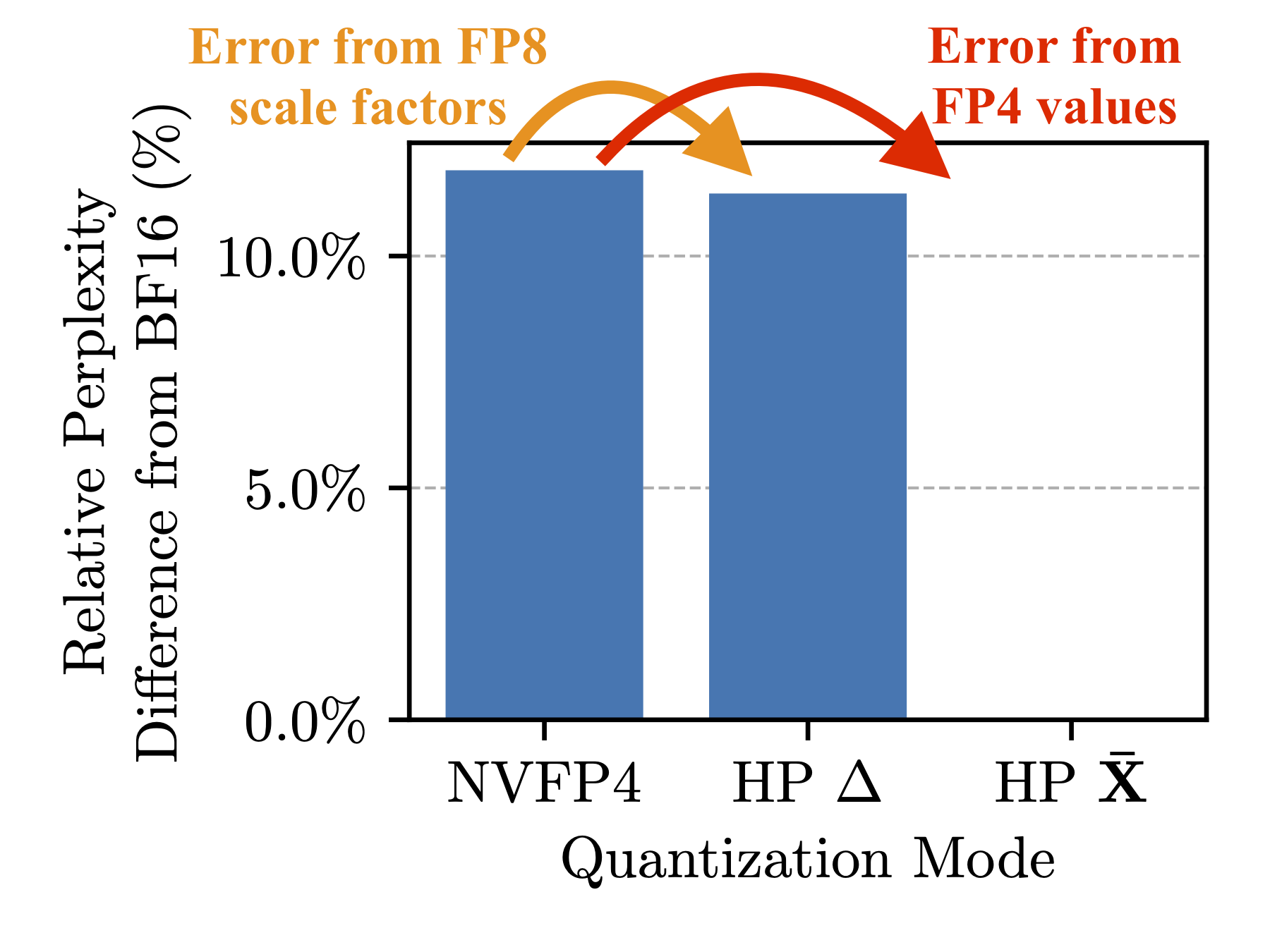}
        \caption{\textbf{Scale Factor Quantization Error has Minimal Effects on Performance.} NVFP4 quantization alongside simulated versions of NVFP4 where either values ($\mathbf{\bar{X}}$) or scale factors ($\Delta$) are kept in high precision (HP). Error from scale factors contributes relatively little to performance loss, but if values are stored in high precision, performance can recover completely.}
        \label{fig:nvfp4-error-types}
    \end{subfigure}
    \hfill
    \begin{subfigure}[t]{0.53\textwidth}
        \centering
        \includegraphics[width=\linewidth]{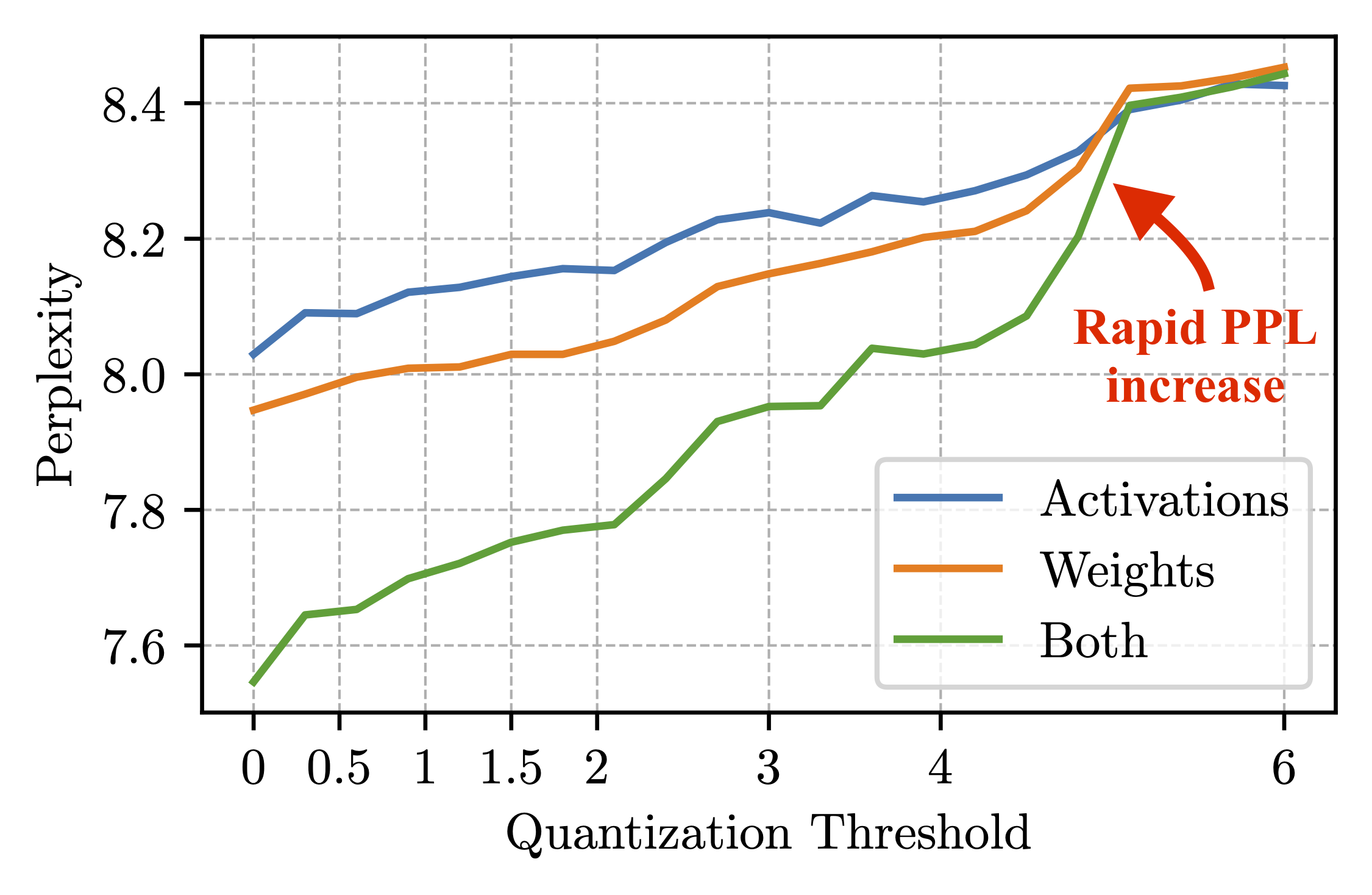}
        \caption{\textbf{NVFP4 Error Comes from Rounding Near-Maximal Values.} Simulated quantization where only values with a magnitude $\geq x$ are quantized to NVFP4. $x=0$, where all values are kept in high precision, is equivalent to BF16, and $x=6$ is equivalent to NVFP4. The steeper slope $x=5$ indicates that error on values quantized around 5, where FP4 has no representable values, are primarily responsible for NVFP4's poor performance.}
        \label{fig:nvfp4-error-thresholds}
    \end{subfigure}
    \caption{Simulated NVFP4 quantization with Llama-3.1-8B evaluated on WikiText-2 word perplexity. To improve NVFP4 performance, we find that we should focus on improving the representation of specific values in each block.}
    \label{fig:nvfp4-error-analysis}
\end{figure}

To better understand the effects of NVFP4 quantization on LLM performance, it would be helpful to understand the downstream effects of each type of NVFP4 quantization error.
In \cref{fig:nvfp4-error-types}, we show how the performance of Llama-3.1-8B is affected when quantized to NVFP4, and when each type of error is mitigated by keeping either scale factors or values in high precision.
Following from standard practice with PTQ, we keep sensitive layers and operations, including embedding layers, the LM head, and Attention calculations in high precision.
We find that error from scale factors has a minimal effect on model performance, and that NVFP4 performance degradation can be entirely attributed to error introduced by casting values to FP4.
When this source of error is mitigated, downstream performance recovers completely.

Next, we simulate how model performance is affected when only some values are casted to FP4.
Specifically, we only cast scaled values to FP4 if their absolute value is above a threshold $\alpha$, measure downstream model performance, and plot our results in \cref{fig:nvfp4-error-thresholds}.
We find that performance degrades steadily up until values greater than 4 are quantized, and then rapidly degrades after that, indicating that error due to rounding values around 5 is most responsible for performance degradation.
Smaller spikes can also be observed at scaled values around 2.5 and 3.5.
This finding aligns with \cref{fig:scaled-to-6}, which shows that these three values suffer from large amounts of error once quantized to FP4.

\begin{figure}
    \centering
    \includegraphics[width=0.9\linewidth]{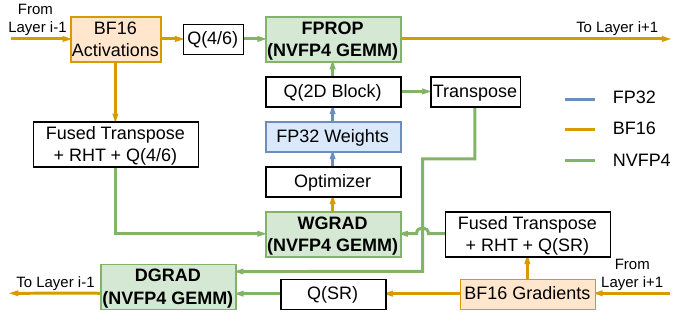}
    \caption{\textbf{Computational flow of an NVFP4 quantized linear layer trained with Four Over Six.} All matrix multiplications (\textbf{FPROP}, \textbf{DGRAD}, \textbf{WGRAD}) are performed in NVFP4, while model weights are stored in FP32, and activations and gradients are stored in BF16. \textbf{Q(4/6)} denotes our method where blocks are scaled to either 4 or 6 based on the distribution of values. \textbf{SR} denotes Stochastic Rounding, and \textbf{RHT} denotes the Random Hadamard Transform. Blue paths represent FP32 data flow; orange paths represent BF16; green paths represent NVFP4.}
    \label{fig:linear}
\end{figure}

This has a large impact on model training: in order to benefit from hardware support, both operands of an NVFP4 matrix multiplication must be quantized to NVFP4.
In \cref{fig:linear}, we show the computational flow of an NVFP4 linear layer, in which weights, activations, and gradients during training are all quantized to NVFP4 before being used.
Following from recent work on training with FP4~\cite{nvidia_pretraining_2025,chmiel_fp4_2025,tseng_training_2025,castro_quartet_2025}, we perform stochastic rounding on gradients to reduce quantization bias, and we perform a random Hadamard transform on the inputs to the weight gradient calculation~\cite{nvidia_pretraining_2025}.
Our experiments above indicate that near-maximal values are responsible for performance degradation during post-training quantization, but it is likely that this explains much of the performance loss during training as well.

\section{Adaptive Block Scaling with Four Over Six}
\label{sec:method}

In the previous section, we found that in NVFP4 quantization, scaled values around 5 in weights and activations are responsible for a large portion of the performance degradation in quantized models, since these values cannot be represented accurately by FP4.
To improve the representation of these values, in this section we introduce \textit{Four Over Six (4/6)}, a method for improving the representation of these near-maximal values in NVFP4 blocks.

Rather than scaling all blocks by the same value $M^\text{FP4}$ as is done in \cref{eqn:fp4-quantization}, we find that changing this scale for some blocks can improve their representation of near-maximal values.
This value dictates the range to which values are quantized to: by setting it to 6, values are spread out over the range of all FP4 values from -6 to 6.
However, we find that for some blocks, it is better to instead set this value to 4, spreading out values over the range of -4 to 4.
By giving up the ability to represent -6 and 6, we allow NVFP4 to represent near-maximal values more accurately.
For example, when scaling to 6, the quantized value 4 represents 66.6\% of the block's maximum value, and no values can be represented between 66.6\% and 100\%.
When a block is instead scaled to 4, the quantized value 3 represents 75\% of the block's maximum value, offering a better ability to represent values in some blocks, such as the samples shown in \cref{tab:sample-quantization}.
Other possible scales, such as 2 or 3, only offer subsets of values that can be represented with 4 or 6.

\begin{table}
    \centering
    \begin{tabular}{rlcc}
        \toprule
           & Pseudocode                                                                                                               & $\mathbf{X}$ = [10, 20, 30, 40]                        & $\mathbf{X}$ = [15, 30, 120, 180]                        \\
        \midrule
        1  & $\Delta^{(6)}$ = max($|\mathbf{X}|$) $\div$ 6                                                                            & 6.67                                                   & 30                                                       \\
        2  & $\Delta^{(6)}$ = \Circled{fp8\_e4m3($\Delta^{(6)}$)}                                                                     & \Circled{6.5}                                          & 30                                                       \\
        3  & $\mathbf{\bar{X}}_i^{(6)}$ = $\mathbf{X}_i$ $\div$ $\Delta^{(6)}$                                                        & [1.54, 3.08, 4.62, 6.15]                               & [0.5, 1, 4, 6]                                           \\
        4  & $\mathbf{\bar{X}}_i^{(6)}$ = \Circled{fp4\_e2m1($\mathbf{\bar{X}}_i^{(6)}$)}                                             & [\Circled{1.5}, \Circled{3}, \Circled{4}, \Circled{6}] & [0.5, 1, 4, 6]                                           \\
        5  & $\mathbf{D}_i^{(6)}$ = $\mathbf{\bar{X}}_i^{(6)}$ $\times$ $\Delta^{(6)}$                                                & [9.75, 19.5, 26, 39]                                   & [15, 30, 120, 180]                                       \\
        6  & $\text{E}^{(6)}$ = $\frac{1}{n}\sum_i^n(\mathbf{D}_i^{(6)}-\mathbf{X}_i)^2$                                              & 4.33                                                   & 0                                                        \\
        \midrule
        7  & $\Delta^{(4)}$ = max($|\mathbf{X}|$) $\div$ 4                                                                            & 10                                                     & 45                                                       \\
        8  & $\Delta^{(4)}$ = \Circled{fp8\_e4m3($\Delta^{(4)}$)}                                                                     & 10                                                     & \Circled{44}                                             \\
        9  & $\mathbf{\bar{X}}_i^{(4)}$ = $\mathbf{X}_i$ $\div$ $\Delta^{(4)}$                                                        & [1, 2, 3, 4]                                           & [0.34, 0.68, 2.73, 4.09]                                 \\
        10 & $\mathbf{\bar{X}}_i^{(4)}$ = \Circled{fp4\_e2m1($\mathbf{\bar{X}}_i^{(4)}$)}                                             & [1, 2, 3, 4]                                           & [\Circled{0.5}, \Circled{0.5}, \Circled{3}, \Circled{4}] \\
        11 & $\mathbf{D}_i^{(4)}$ = $\mathbf{\bar{X}}_i^{(4)}$ $\times$ $\Delta^{(4)}$                                                & [10, 20, 30, 40]                                       & [22, 22, 136, 176]                                       \\
        12 & $\text{E}^{(4)}$ = $\frac{1}{n}\sum_i^n(\mathbf{D}_i^{(4)}-\mathbf{X}_i)^2$                                              & 0                                                      & 96.25                                                    \\
        \midrule
        13 & $\Delta$ = $\Delta^{(4)}$ if $\text{E}^{(4)}$ < $\text{E}^{(6)}$ else $\Delta^{(6)}$                                     & 10                                                     & 30                                                       \\
        14 & $\mathbf{\bar{X}}_i$ = $\mathbf{\bar{X}}_i^{(4)}$ if $\text{E}^{(4)}$ < $\text{E}^{(6)}$ else $\mathbf{\bar{X}}_i^{(6)}$ & [1, 2, 3, 4]                                           & [0.5, 1, 4, 6]                                           \\
        \bottomrule
    \end{tabular}
    \vspace{0.2cm}
    \caption{We highlight how for these two sample blocks, our procedure may choose to scale the block using either 4 or 6. The standard NVFP4 quantization algorithm ends on line 4, returning $\Delta^{(6)}$ and $\mathbf{\bar{X}}^{(6)}$. Operations that introduce error, and values that suffer from this error, are highlighted in orange.}
    \label{tab:sample-quantization}
\end{table}

\subsection{Scale Selection Rules}
\label{sec:scale_selection_rules}

{
\begin{table}
    \centering
    \small
    \begin{tabular}{l|ccc|ccc}
        \toprule
                    & \multicolumn{3}{|c}{Llama 3} & \multicolumn{3}{c}{Qwen 3}                                                                   \\
                    & 1B                           & 8B                         & 70B           & 1.7B           & 8B             & 32B           \\
        \midrule
        BF16        & 11.98                        & 7.54                       & 2.86          & 21.06          & 12.22          & 9.34          \\
        \midrule
        MXFP4       & 17.67                        & 9.66                       & 5.53          & 26.95          & 14.08          & 11.92         \\
        NVFP4 (M=6) & \textbf{14.27}               & \textbf{8.43}              & \textbf{4.00} & \textbf{23.06} & \textbf{12.68} & \textbf{9.85} \\
        NVFP4 (M=4) & 14.75                        & 8.63                       & 4.48          & 24.43          & 12.98          & 10.57         \\
        \bottomrule
    \end{tabular}
    \vspace{0.2cm}
    \caption{\textbf{Scaling Blocks to 4 Is Not Always Better.} WikiText-2 word perplexity for models quantized with W4A4 PTQ. MXFP4 and baseline NVFP4 quantization ($M=6$) suffer from performance degradation. Scaling all blocks with $M=4$ further degrades overall performance.}
    \label{tab:mxfp4-vs-nvfp4}
\end{table}
}

{
\setlength{\tabcolsep}{4pt}
\begin{table}
    \centering
    \small
    \begin{tabular}{l|ccc|ccc|ccc|ccc}
        \toprule
                        & \multicolumn{6}{|c}{WikiText-2 Word PPL} & \multicolumn{6}{|c}{C4 Word PPL}                                                                                                                                                                                                         \\
                        & \multicolumn{3}{|c}{Llama 3}    & \multicolumn{3}{|c}{Qwen 3} & \multicolumn{3}{|c}{Llama 3} & \multicolumn{3}{|c}{Qwen 3}                                                                                                                                        \\
                        & 1B                              & 8B                          & 70B                          & 1.7B                        & 8B             & 32B           & 1B             & 8B             & 70B            & 1.7B           & 8B             & 32B            \\
        \toprule
        BF16            & 11.98                           & 7.54                        & 2.86                         & 21.06                       & 12.22          & 9.34          & 28.54          & 18.08          & 12.52          & 58.76          & 36.35          & 26.12          \\
        \midrule
        RTN             & 14.27                           & 8.43                        & 4.00                         & \textbf{23.06}              & 12.68          & 9.85          & 36.19          & 20.83          & 14.16          & 65.54          & 37.91          & \textbf{27.54} \\
        + 4/6 (MSE)     & \textbf{13.84}                  & \textbf{8.30}               & \textbf{3.83}                & 23.60                       & \textbf{12.56} & 9.84          & \textbf{35.09} & \textbf{20.48} & 13.95          & 66.32          & \textbf{37.32} & 27.67          \\
        + 4/6 (L1)      & 13.94                           & 8.33                        & 3.86                         & 23.45                       & 12.63          & \textbf{9.82} & 35.32          & 20.56          & \textbf{13.94} & 66.41          & 37.53          & 27.56          \\
        + 4/6 (Abs-Max) & 14.06                           & 8.36                        & 4.39                         & 23.32                       & 12.86          & 9.97          & 35.43          & 20.68          & 14.31          & \textbf{65.39} & 37.93          & 28.21          \\
        \bottomrule
    \end{tabular}
    \vspace{0.2cm}
    \caption{\textbf{Selecting Block Scales Using MSE Works Best Overall.} After a block is quantized to NVFP4 with $N=4$ and $N=6$, there are several ways to pick which quantized version is better. We find that using mean squared quantization error generally works best.}
    \label{tab:scale-selection-rules}
\end{table}
}

Scaling blocks to 4 can reduce quantization error on some blocks, but not on all blocks.
This is demonstrated by \cref{tab:sample-quantization}, in which one block is better represented when scaled to 6, and another block is better represented when scaled to 4.
In \cref{tab:mxfp4-vs-nvfp4}, we evaluate several Llama~\cite{grattafiori_llama_2024} and Qwen~\cite{yang_qwen3_2025} models when quantized to NVFP4 when all blocks are scaled to 4, and with the standard formulation in which all blocks are scaled to 6.
We find that despite its ability to represent near-maximal values more accurately, scaling all blocks to 4 results in worse performance than scaling all blocks to 6 as is done in standard NVFP4 quantization, likely due to the significantly larger range of values that can be expressed with a scale of 6 ($\frac{6}{0.5}$ vs. $\frac{4}{0.5}$, a 50\% increase).
Instead, blocks need to be scaled adaptively.

We find that it is difficult to identify which scale is best without access to a block's quantized values.
To implement Four Over Six, we quantize each block twice: once with a scale of 6, and again with a scale of 4, and then compare the quantized values $\bar{\mathbf{X}}^{(4)}$ and $\bar{\mathbf{X}}^{(6)}$ to the original values $\mathbf{X}$.
With access to these values, we implement three different scale selection rules: selecting the quantized values with the smallest maximum error (i.e., $\max_i(\mathbf{\bar{X}}^{(4)}_i-\mathbf{X}_i)$ vs. $\max_i(\mathbf{\bar{X}}^{(6)}_i-\mathbf{X}_i)$), with the lower mean absolute error (MAE), and with the lower mean squared error (MSE).
In \cref{tab:scale-selection-rules}, we show the effects of these three rules on model performance.
We find that each rule sometimes outperforms standard NVFP4 quantization, but that selecting a block's scale based on quantization MSE works best in most cases.
In the remainder of this work, we adopt the MSE scale selection rule for all post-training quantization experiments.

Finally, we make one modification to the computation of the tensor scale $\alpha$ (\cref{eqn:fp4-quantization-tensor-scale}) when quantizing to NVFP4 with 4/6.
When $M^\text{FP4} \times M^\text{FP8}$ is used to compute the tensor scale, it ensures that all quantized values will be less than $6 \times 448$.
However, this makes it impossible to select a scale of 4 for the blocks that contain a tensor's largest values, because the block's scale would need to be $448 \times \frac{6}{4}=672$, which would overflow since 448 is the maximum value that can be represented by E4M3.
As a result, when computing the tensor scale, we replace $M^\text{FP8}$ to 256 in \cref{eqn:fp4-quantization-tensor-scale}, since 256 is the largest E4M3 that can be multiplied by $\frac{6}{4}$ and represented without error in E4M3, as 384.

\subsection{Implementation}

We find that Four Over Six can be implemented efficiently using PTX instructions supported by NVIDIA Blackwell GPUs.
Specifically, we use the \texttt{cvt} family of instructions to perform quantization into the packed FP4 format, and then also dequantization from FP4 to FP16, which is needed to calculate error as is done in lines 6 and 12 of \cref{tab:sample-quantization}.
To maintain high performance, we implement Four Over Six in a CUDA kernel where all quantized values, dequantized values, and errors are kept in the register file.
As a result, we observe that the overhead introduced to the NVFP4 quantization kernel by Four Over Six is under 15\%, and we expect that we will be able to reduce this overhead further with more optimization.
\section{Evaluation}

In this section, we evaluate how Four Over Six affects the performance of models quantized to NVFP4 during pre-training and post-training quantization.
We find that the addition of Four Over Six leads to improved performance during pre-training, and improves downstream performance of already-trained Llama~\cite{grattafiori_llama_2024} and Qwen~\cite{yang_qwen3_2025} models across several downstream tasks.

\subsection{Pre-Training}

\begin{figure}
    \centering
    \includegraphics[width=\linewidth]{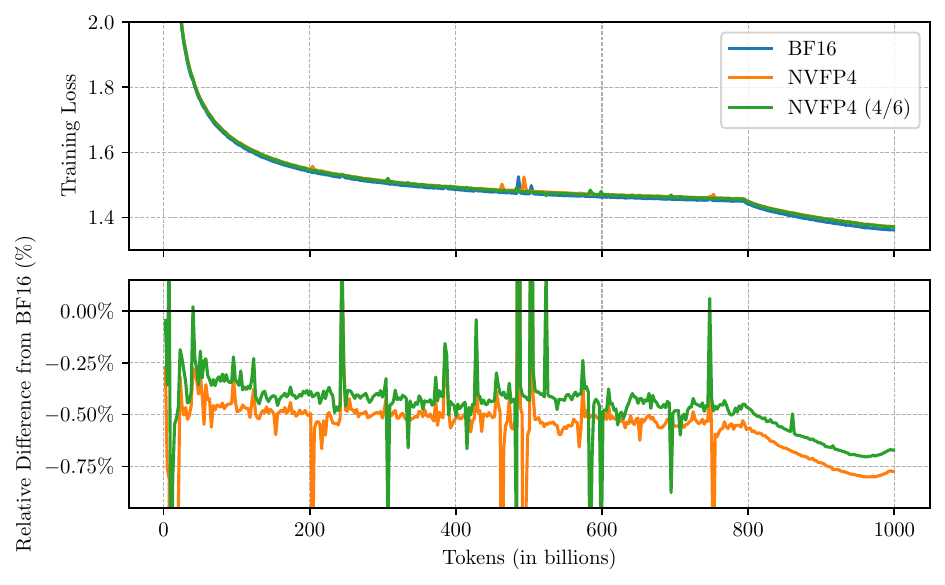}
    \caption{\textbf{Four Over Six Improves Performance During NVFP4 Pre-Training.} Training loss curves comparing BF16, NVFP4, and NVFP4 with 4/6 for our Mixture-of-Experts hybrid Mamba-Transformer model architecture. Adding 4/6 keeps NVFP4 training loss closer to BF16.}
    \label{fig:pretraining}
\end{figure}

\cref{fig:pretraining} shows our main results during pre-training experiments with the Nemotron 3 Nano 30B-A3B model architecture~\cite{nvidia_nemotron_2025}.
We find that 4/6 improves performance compared to the NVFP4 baseline, bringing training loss 13.0\% closer to the BF16 baseline.
Notably, we find that while 4/6 improves performance when used to quantize activations, weights, and gradients, comparable performance can be obtained when 4/6 is only used during activation quantization (\cref{fig:pretraining_tensor_specific}).
Contrary to our findings from post-training quantization experiments, we find that performance is best when using mean absolute error (MAE) to select scale factors (\cref{fig:pretraining_scale_rules}).
In addition to reduced training loss, these two design choices are preferable as they keep the computational overhead introduced by 4/6 to a minimum.


We train our models on a high-quality curated and synthetic data distribution of 1 trillion tokens based on \cite{nvidia_nvidia_2025} using the AdamW optimizer~\cite{loshchilov_decoupled_2019} with $\beta_1=0.9$, $\beta_2=0.95$, weight decay of $0.1$, gradient clipping at $1.0$, a sequence length of 8192, and a global batch size of 3072.
We use a Warmup-Stable-Decay~\cite{hu_minicpm_2024} learning rate schedule with a constant learning rate of $10^{-3}$ which decays to $10^{-5}$ over the last 20\% of training, and we train all models using 384 NVIDIA B200 GPUs.

Following from the current state-of-the-art NVFP4 pre-training recipe~\cite{nvidia_pretraining_2025}, we perform stochastic rounding on gradients, apply a random Hadamard transform to both inputs of the weight gradient calculation, and perform 2D block quantization on weight matrices, as outlined in \cref{fig:linear}.
All NVFP4 matrix multiplications accumulate in FP32 and output in BF16, and model weights are stored in FP32.
Following from \cite{nvidia_pretraining_2025}, Attention components, the output projection head, normalization layers, and non-linearities are kept in high precision, either BF16 or FP32.
We also keep the output projection layer of Mamba-2 blocks in MXFP8.

\subsection{Post-Training Quantization}

We also evaluate the ability of 4/6 to improve NVFP4 post-training quantization (PTQ) accuracy.
While 4/6 can be used on its own, 4/6 is a general method that modifies the underlying NVFP4 quantization algorithm, allowing it to be easily combined with existing PTQ methods such as GPTQ~\cite{frantar_gptq_2023,egiazarian_bridging_2025}, AWQ~\cite{lin_awq_2024}, and SmoothQuant~\cite{xiao_smoothquant_2024}.
When evaluating Llama 3 and Qwen3 models on WikiText-2 and C4 word perplexity, we find that 4/6 improves performance in the vast majority of cases, as shown in \cref{tab:eval-perplexity}.
When combined with AWQ and SmoothQuant, 4/6 improves performance on these metrics for all models tested, bringing word perplexity 19.9\% and 5.3\% closer to BF16 model performance respectively.
We find that 4/6 reduces the performance of models quantized with GPTQ, increasing the gap between NVFP4 and BF16 word perplexity by an average of 34.6\%.
AWQ with 4/6 performs best overall, with an average WikiText-2 word perplexity of 11.58 and an average C4 word perplexity of 32.36 across all models.

To evaluate GPTQ, we use the FP-Quant implementation available on GitHub~\cite{egiazarian_bridging_2025} and load quantized models with either the built-in FP4 linear layers or our version which performs quantization with 4/6.
Modifying the GPTQ optimization process in a way that incorporates Four Over Six is likely to deliver performance improvements in future work.

While perplexity is often considered a more stable metric for evaluating quantized models~\cite{dettmers_case_2023}, we also evaluate these quantized models on several downstream tasks including BoolQ~\cite{clark_boolq_2019}, ARC-Easy and ARC-Challenge~\cite{clark_think_2018}, and HellaSwag~\cite{zellers_hellaswag_2019}.
We report normalized accuracy for ARC-Easy, ARC-Challenge, and HellaSwag in order to reduce differences due to tokenization when comparing across different models.
We find that 4/6 improves performance across most PTQ methods and tasks, with average task performance improved in nearly all cases, as shown in \cref{tab:eval-tasks-llama3} for Llama 3 models and \cref{tab:eval-tasks-qwen3} for Qwen3 models.

{
    \setlength{\tabcolsep}{4pt}
    \begin{table}
        \centering
        \small
        \begin{tabular}{l|ccc|ccc|ccc|ccc}
            \toprule
                        & \multicolumn{6}{|c}{WikiText-2 Word PPL ($\downarrow$)} & \multicolumn{6}{|c}{C4 Word PPL ($\downarrow$)}                                                                                                                                                                                                         \\
                        & \multicolumn{3}{|c}{Llama 3}    & \multicolumn{3}{|c}{Qwen 3} & \multicolumn{3}{|c}{Llama 3} & \multicolumn{3}{|c}{Qwen 3}                                                                                                                                        \\
                        & 1B                              & 8B                          & 70B                          & 1.7B                        & 8B             & 32B           & 1B             & 8B             & 70B            & 1.7B           & 8B             & 32B            \\
            \toprule
            BF16        & 11.98                           & 7.54                        & 2.86                         & 21.06                       & 12.22          & 9.34          & 28.54          & 18.08          & 12.52          & 58.76          & 36.35          & 26.12          \\
            \midrule
            RTN         & 14.27                           & 8.43                        & 4.00                         & \textbf{23.06}              & 12.68          & 9.85          & 36.19          & 20.83          & 14.16          & \textbf{65.54} & 37.91          & \textbf{27.54} \\
            + 4/6       & \textbf{13.84}                  & \textbf{8.30}               & \textbf{3.83}                & 23.60                       & \textbf{12.56} & \textbf{9.84} & \textbf{35.09} & \textbf{20.48} & \textbf{13.95} & 66.32          & \textbf{37.32} & 27.67          \\
            \midrule
            GPTQ        & 13.73                           & 8.33                        & --                           & \textbf{21.48}              & \textbf{12.50} & 9.67          & 35.65          & 20.98          & --             & \textbf{63.33} & \textbf{37.14} & 27.17          \\
            + 4/6       & \textbf{13.67}                  & \textbf{8.30}               & --                           & 22.70                       & 12.65          & \textbf{9.66} & \textbf{35.55} & \textbf{20.89} & --             & 63.81          & 37.25          & \textbf{27.09} \\
            \midrule
            AWQ         & 14.04                           & 8.33                        & 3.86                         & 22.20                       & 12.68          & 9.69          & 35.56          & 20.69          & 13.58          & 62.50          & 37.51          & 27.09          \\
            + 4/6       & \textbf{13.67}                  & \textbf{8.24}               & \textbf{3.71}                & \textbf{21.67}              & \textbf{12.57} & \textbf{9.64} & \textbf{34.55} & \textbf{20.34} & \textbf{13.41} & \textbf{61.78} & \textbf{37.14} & \textbf{26.95} \\
            \midrule
            SmoothQuant & 14.17                           & 8.38                        & 3.86                         & 21.99                       & 12.64          & 9.65          &         35.61       &    20.80            & --             &           62.33     &       37.67         & --             \\
            + 4/6       & \textbf{14.03}                  & \textbf{8.32}               & \textbf{3.80}                & \textbf{21.97}              & \textbf{12.62} & \textbf{9.63} &           \textbf{35.20}     &   \textbf{20.61}             & --             &  \textbf{62.19}              &    \textbf{37.62}            & --             \\
            \bottomrule
        \end{tabular}
        \vspace{0.2cm}
        \caption{\textbf{4/6 Can Improve the Performance of Existing PTQ Methods.} We find that 4/6 uniformly improves perplexity metrics when combined with AWQ and SmoothQuant, and can also improve the performance of RTN (round-to-nearest) quantization and GPTQ in most cases.}
        \label{tab:eval-perplexity}
    \end{table}
}

\begin{table}
    \centering
    \begin{tabular}{l|cc|cc|cc|cc|cc}
        \toprule
        \small
                    & \multicolumn{2}{|c}{BoolQ ($\uparrow$)} & \multicolumn{2}{|c}{Arc-E ($\uparrow$)} & \multicolumn{2}{|c}{Arc-C ($\uparrow$)} & \multicolumn{2}{|c}{HellaSwag ($\uparrow$)} & \multicolumn{2}{|c}{\textbf{Average ($\uparrow$)}}                                                                                 \\
                    & 1B                         & 8B                         & 1B                         & 8B                             & 1B                                    & 8B            & 1B            & 8B            & 1B            & 8B            \\
        \toprule
        BF16        & 63.7                       & 83.2                       & 61.8                       & 82.5                           & 37.0                                  & 55.0          & 64.3          & 79.3          & 56.7          & 75.0          \\
        \midrule
        RTN         & \textbf{58.6}              & 80.2                       & \textbf{57.3}              & 77.5                           & \textbf{34.3}                         & 52.5          & 60.0          & 77.7          & \textbf{52.3} & 72.0          \\
        + 4/6       & 57.7                       & \textbf{80.9}              & 56.9                       & \textbf{80.2}                  & 33.0                                  & \textbf{52.6} & 60.0          & \textbf{78.0} & 51.9          & \textbf{72.2} \\
        \midrule
        GPTQ        & \textbf{61.1}              & 80.3                       & 57.3                       & 78.3                           & 33.1                                  & \textbf{53.9} & 60.6          & 77.0          & 53.0          & 72.4          \\
        + 4/6       & 60.2                       & \textbf{81.4}              & \textbf{57.5}              & \textbf{78.7}                  & \textbf{33.8}                         & 53.0          & \textbf{60.7} & \textbf{77.4} & \textbf{53.1} & \textbf{72.6} \\
        \midrule
        AWQ         & 59.8                       & \textbf{81.3}              & 58.0                       & 78.4                           & 34.2                                  & 51.7          & 60.9          & 77.5          & 53.2          & 72.2          \\
        + 4/6       & \textbf{61.0}              & 80.4                       & \textbf{58.8}              & \textbf{80.2}                  & \textbf{35.5}                         & \textbf{53.6} & \textbf{61.2} & \textbf{78.2} & \textbf{54.1} & \textbf{73.1} \\
        \midrule
        SmoothQuant & 61.1                       & \textbf{81.1}              & 57.4                       & 77.6                           & 35.5                                  & \textbf{52.7} & 60.7          & 77.6          & 53.7          & 72.3          \\
        + 4/6       & \textbf{61.6}              & 80.4                       & \textbf{58.0}              & \textbf{78.6}                  & \textbf{35.6}                         & 52.2          & \textbf{61.6} & \textbf{80.4} & \textbf{54.2} & \textbf{72.9} \\
        \bottomrule
    \end{tabular}
    \vspace{0.2cm}
    \caption{Downstream task performance of Llama-3.2-1B and Llama-3.1-8B when quantized using various PTQ methods and 4/6. 4/6 improves average task performance in nearly all cases.}
    \label{tab:eval-tasks-llama3}
\end{table}

\begin{table}
    \centering
    \begin{tabular}{l|cc|cc|cc|cc|cc}
        \toprule
        \small
                    & \multicolumn{2}{|c}{BoolQ ($\uparrow$)} & \multicolumn{2}{|c}{Arc-E ($\uparrow$)} & \multicolumn{2}{|c}{Arc-C ($\uparrow$)} & \multicolumn{2}{|c}{HellaSwag ($\uparrow$)} & \multicolumn{2}{|c}{\textbf{Average ($\uparrow$)}}                                                                                 \\
                    & 1.7B                       & 8B                         & 1.7B                       & 8B                             & 1.7B                                  & 8B            & 1.7B          & 8B            & 1.7B          & 8B            \\
        \toprule
        BF16        & 77.6                       & 86.6                       & 70.2                       & 80.9                           & 43.0                                  & 56.7          & 60.4          & 74.9          & 62.8          & 74.8          \\
        \midrule
        RTN         & 73.1                       & 85.5                       & 59.9                       & \textbf{78.6}                  & 35.3                                  & \textbf{54.1} & 58.0          & 73.2          & 56.6          & 72.3          \\
        + 4/6       & \textbf{76.3}              & \textbf{85.8}              & \textbf{66.3}              & 78.5                           & \textbf{38.7}                         & 53.7          & 58.0          & \textbf{73.7} & \textbf{59.8} & \textbf{72.9} \\
        \midrule
        GPTQ        & \textbf{74.6}              & 86.4                       & \textbf{60.9}              & 79.8                           & 36.3                                  & 54.3          & \textbf{57.2} & 73.2          & \textbf{57.3} & 73.4          \\
        + 4/6       & 73.8                       & \textbf{86.5}              & 59.8                       & \textbf{80.8}                  & \textbf{37.3}                         & \textbf{54.9} & 57.1          & 73.2          & 57.0          & \textbf{73.9} \\
        \midrule
        AWQ         & 72.8                       & 86.5                       & 58.0                       & \textbf{78.4}                  & 36.3                                  & 55.2          & 57.8          & 73.6          & 56.2          & 73.4          \\
        + 4/6       & \textbf{75.9}              & 86.5                       & \textbf{63.8}              & 78.1                           & \textbf{39.2}                         & \textbf{55.8} & \textbf{57.9} & 73.6          & \textbf{59.2} & \textbf{73.5} \\
        \midrule
        SmoothQuant & \textbf{74.8}              & \textbf{86.4}              & 61.2                       & 78.1                           & 37.2                                  & \textbf{55.2} & 57.8          & 73.3          & 57.8          & 73.2          \\
        + 4/6       & 74.5                       & 85.9                       & \textbf{61.7}              & \textbf{78.4}                  & \textbf{40.4}                         & 54.8          & \textbf{57.9} & \textbf{73.5} & \textbf{58.6} & 73.2          \\
        \bottomrule
    \end{tabular}
    \vspace{0.2cm}
    \caption{Downstream task performance of Qwen3-1.7B and Qwen3-8B when quantized using various PTQ methods and 4/6. 4/6 improves or matches average task performance in nearly all cases.}
    \label{tab:eval-tasks-qwen3}
\end{table}
\section{Discussion}

\subsection{Outliers}

Modern training techniques often aim to mitigate outliers in weights and activations, since they can often degrade model performance, introduce instability during training, and make models harder to compress~\cite{an_systematic_2025,xiao_efficient_2024,bondarenko_quantizable_2023,nrusimha_mitigating_2024}.
However, by introducing a separate scale factor for every $m$ values, block-scaled floating point formats such as MXFP4 ($m=32$) and NVFP4 ($m=16$) are able to represent outliers with almost no error.
As a result, most of the quantization error in these formats comes from near-maximal values, as demonstrated in \cref{sec:quantizing-llms-to-nvfp4}.
While we are able to adapt NVFP4 to represent these values with less error, some models with very few outliers may benefit further from formats with even more uniform quantization error, such as INT4~\cite{chen_int_2025}.
However, NVFP4, especially once combined with 4/6, has empirically proven better at quantizing models in most real-world cases.

\subsection{Limitations}

While Four Over Six could theoretically be applied to other block-scaled low-precision FP4 formats, we focus on NVFP4 in this work because Four Over Six would not work with MXFP4.
Note that the ability to scale blocks to either 4 or 6 requires a minimum amount of precision to be present in scale factors: a block scaled to 4 requires a scale factor that is 50\% larger than a block scaled to 6.
This is possible to represent with FP8 E4M3, the numerical format used by NVFP4 to represent scale factors.
However, this is not possible in MXFP4~\cite{rouhani_microscaling_2023} scale factors, which are saved in FP8 E8M0, a format in which each representable value is a factor of 2 away from the previous or next representable value.
Future block-scaled floating point formats may benefit from 4/6-style Adaptive Block Scaling, however the benefits fade quickly as the precision used to store values increases.
\section{Related Work}
\label{sec:related_work}

\textbf{Quantization} has seen widespread adoption in LLMs due to its ability to reduce model size and accelerate inference~\cite{han_deep_2016}.
Most methods aim to mitigate outliers in order to reduce the dynamic range of tensors that need to be quantized.
This is often done with per-channel smoothing factors~\cite{lin_awq_2024,xiao_smoothquant_2024}, second-order information~\cite{frantar_gptq_2023}, or rotations~\cite{liu_spinquant_2025,ashkboos_quarot_2024}.
Most of these works were developed with other numerical formats in mind, such as INT4, which does not have dedicated support on newer hardware accelerators.

\textbf{Block-scaled quantization} formats have grown in popularity due to the increased precision they provide.
For example, the smallest nonzero value that can be represented in FP4 is 0.5, and the largest is 6, meaning a tensor quantized to FP4 should ideally have a dynamic range of 12, far too small for many practical applications.
Block-scaled formats such as MXFP4~\cite{rouhani_microscaling_2023} and NVFP4~\cite{nvidia_pretraining_2025}, on the other hand, store a higher-precision scaling factor alongside blocks of FP4 values, allowing for different blocks in the same tensor to have vastly different scales.
In the case of MXFP4, every 32 FP4 values are accompanied by an 8-bit E8M0 scaling factor, and in the case of NVFP4, every 16 FP4 values are accompanied by an 8-bit E4M3 scaling factor.
Crucially, both of these methods have dedicated hardware support in the most recently-released NVIDIA Blackwell GPUs, offering up to 2x speed improvements over FP8 inputs, and 4x speed improvements over BF16/FP16 inputs on NVIDIA B200 GPUs.
However, despite the benefits these formats provide, maintaining model performance after quantization remains challenging.

\textbf{Low-precision training} with MXFP4 and NVFP4 remains a challenging problem.
In theory, FP4 training should provide two primary benefits: improved training speed, and a natively-quantized FP4 model.
In practice, achieving these goals is difficult.
Many works use stochastic rounding, the random Hadamard transform, or both to mitigate the effects of outliers~\cite{tseng_training_2025,nvidia_pretraining_2025,castro_quartet_2025,wang_optimizing_2025}.
However, each of these operations, in addition to the process of computing scale factors, adds computational overhead.
If the overhead becomes too large, training with block-scaled FP4 formats becomes pointless, as FP8 formats are generally far more accurate and are only 2x slower on NVIDIA B200 GPUs.
Furthermore, many FP4 training works require ``healing'' the model by training in high-precision for some time afterward, adding to the time cost, and resulting in a high-precision model~\cite{chmiel_fp4_2025,nvidia_pretraining_2025}.

\textbf{Post-training quantization} with MXFP4 and NVFP4 is easier, since models can be calibrated offline to recover accuracy.
SpinQuant~\cite{liu_spinquant_2025} and QuaRot~\cite{ashkboos_quarot_2024} rotate model weights to make them easier to quantize.
AWQ~\cite{lin_awq_2024}, GPTQ~\cite{frantar_gptq_2023}, SmoothQuant~\cite{xiao_smoothquant_2024}, and SVDQuant~\cite{li_svdquant_2025} rely on offline calibration.
However, quantizing to MXFP4 and NVFP4 with these methods still generally fails to recover full high-precision performance, as shown in \cref{tab:mxfp4-vs-nvfp4}.
\section{Conclusion}

In this work, we introduced \textit{Four Over Six}, a change to the NVFP4 quantization algorithm that improves quantization accuracy while introducing minimal overhead.
We find that 4/6 improves pre-training performance compared to existing state-of-the-art NVFP4 pre-training recipes, bringing performance closer to high-precision baselines.
When added to existing PTQ methods, we find that 4/6 leads to broad performance improvements across a variety of tasks.
We hope this work inspires future work in NVFP4 quantization.

\begin{ack}
We thank Modal and NVIDIA for access to the B200 nodes needed to run our experiments.
This research is partially supported by Amazon, Hyundai Motor Company, the MIT AI Hardware Program, the MIT-IBM Watson AI Lab, the National Science Foundation, and the National Science Foundation Graduate Research Fellowship under Grant No. 2141064.
Any opinion, findings, and conclusions or recommendations expressed in this material are those of the authors and do not necessarily reflect the views of the National Science Foundation.
\end{ack}

\bibliographystyle{unsrt}
\bibliography{ref_zotero}


\appendix
\clearpage
\section{Training Methodology}

When adding Four Over Six (4/6) to an NVFP4 pre-training recipe, there are four main decisions to make: which scale selection rule to use, how to combine 4/6 with stochastic rounding, whether to reduce the FP32 tensor scale factor, and which tensors should be quantized with 4/6.
In this section, we ablate each of these design decisions in detail.
We find that the best training recipe involves only using 4/6 when quantizing activations, and that this should be done using the mean absolute error scale selection rule and a reduced FP32 tensor scale.

\subsection{Scale Selection Rule}

In \cref{sec:scale_selection_rules}, we describe how 4/6 can be implemented using various scale selection rules.
Here, we test whether mean squared error (MSE) or mean absolute error (MAE) provides better performance.
We evaluate two settings against BF16 and NVFP4 baselines: one in which activations (X), weights (W), and gradients (G) are all quantized with 4/6, and one in which only activations and weights are quantized with 4/6, and gradients are quantized with standard NVFP4 quantization using stochastic rounding.

\begin{figure}[ht]
    \centering
    \includegraphics[width=\linewidth]{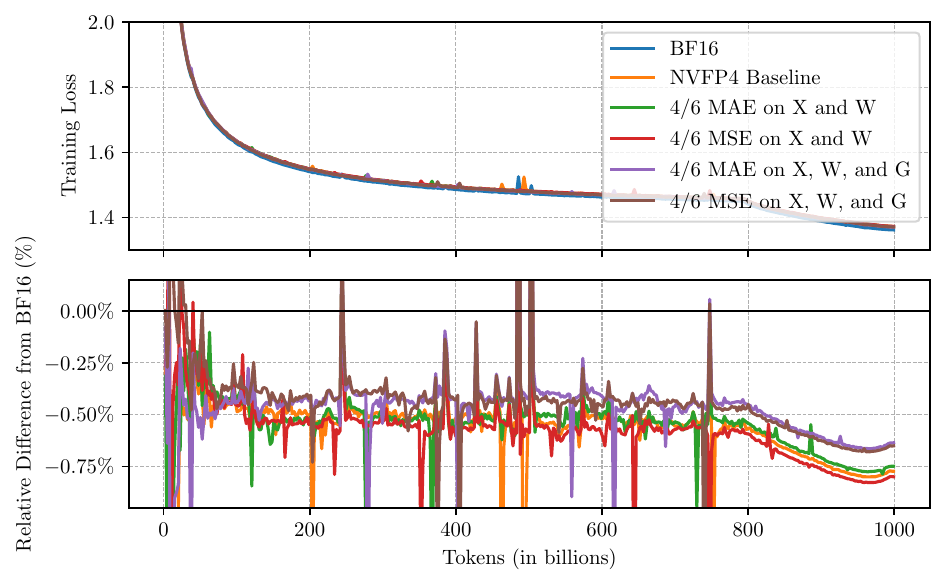}
    \caption{\textbf{Scale Selection with MAE Outperforms MSE For Large-Scale Pre-Training.} Regardless of whether or not gradients are quantized using 4/6, selecting scale factors using mean absolute error results in marginally better performance during pre-training, as measured by training loss.}
    \label{fig:pretraining_scale_rules}
\end{figure}

Results are shown in \cref{fig:pretraining_scale_rules}.
Even though MSE outperforms MAE in most post-training quantization experiments (\cref{sec:scale_selection_rules}), we find that in both pre-training settings, MAE provides better training performance.
In order to isolate the effects of each individual design decision, the reduced FP32 tensor scale described in \cref{sec:scale_selection_rules} is not applied in these experiments.
In all remaining experiments in this section, MAE is used to select scale factors when quantizing with 4/6.

\clearpage
\subsection{Reduced FP32 Tensor Scale}

\begin{figure}[ht]
    \centering
    \includegraphics[width=\linewidth]{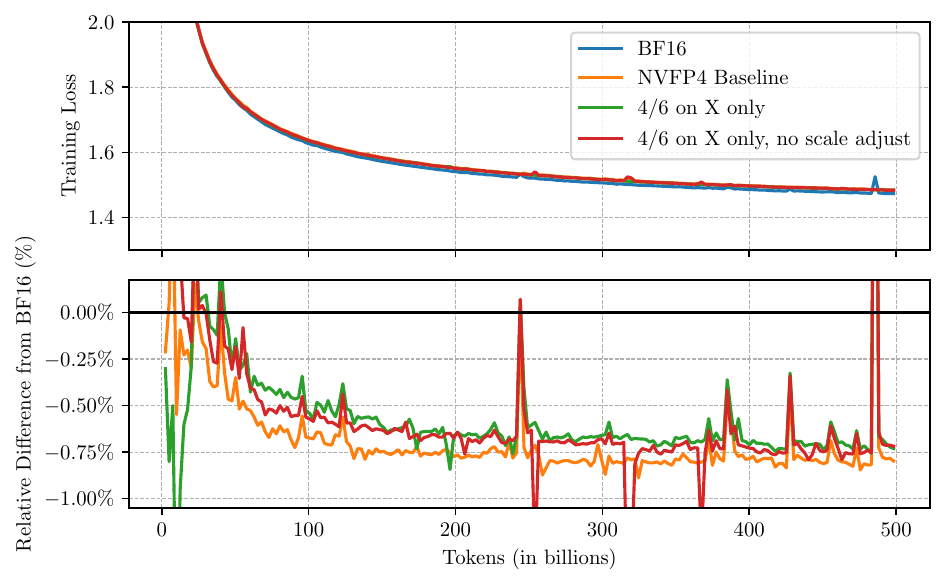}
    \caption{\textbf{Reducing the FP32 Tensor Scale Improves Pre-Training Performance.} When the FP32 global tensor scale is reduced in order to allow blocks with a tensor's largest value to select a maximum value of 4, training performance improves slightly.}
    \label{fig:pretraining_scale_adjust}
\end{figure}

In \cref{sec:scale_selection_rules}, we propose calculating the FP32 global tensor scale using 256 as the maximum FP8 E4M3 value rather than the default of 448, as this allows blocks with a tensor's largest value to have the option to have a largest FP4 value of 4.
In \cref{fig:pretraining_scale_adjust}, we find that this provides a marginal benefit over using the standard tensor scale calculation.
Even though this adjustment only affects a small number of large values, this performance gain may come from the fact that larger activation values can have an outsize impact on model performance.
This adjustment is incorporated into the remaining experiments in this section.

\clearpage
\subsection{Stochastic Rounding}

Current state-of-the-art NVFP4 pre-training recipes involve quantizing gradients with stochastic rounding~\cite{tseng_training_2025,nvidia_pretraining_2025}.
We adopt this practice in our work as well, but this introduces a minor issue: the purpose of stochastic rounding is to eliminate bias introduced during quantization.
However, when 4/6 is used with values that have been rounding stochastically, bias is reintroduced in a way that favors reduced quantization error.
While applying 4/6 to stochastically rounded values works well in practice, we also evaluate two alternative options:

\begin{enumerate}
    \item Using standard NVFP4 quantization with stochastic rounding for gradients without 4/6
    \item Selecting 4 or 6 using round-to-nearest quantization error, then rounding stochastically with the selected scale factor
\end{enumerate}

\begin{figure}[ht]
    \centering
    \includegraphics[width=\linewidth]{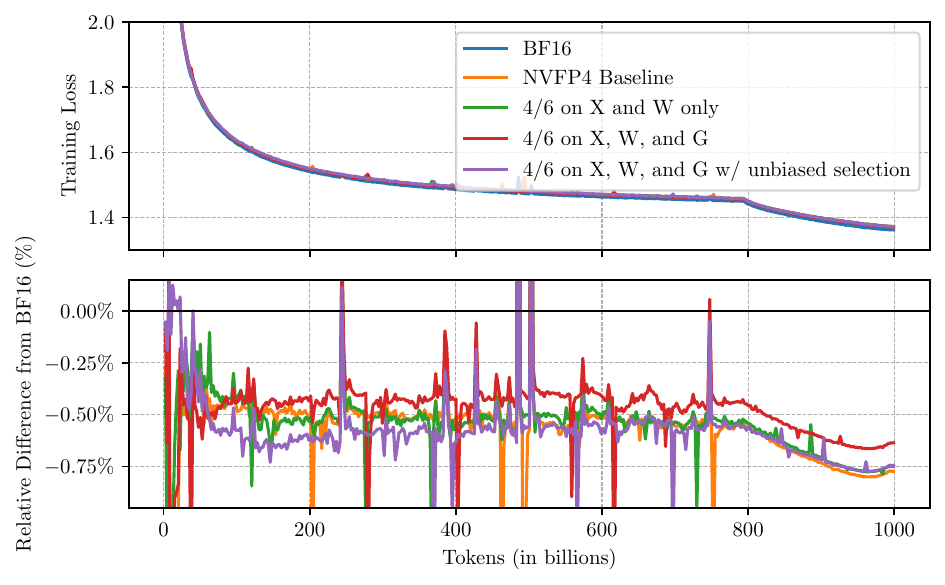}
    \caption{\textbf{If Applied to Gradient Quantization, 4/6 Should Be Used With Biased Selection.} Even though alternative formulations, such as refraining from the use of 4/6 with gradients, or an unbiased selection mechanism, can reduce quantization bias, we find that the default 4/6 formulation yields the best performance during pre-training.}
    \label{fig:pretraining_gradients}
\end{figure}

Both of these options allow for unbiased gradient estimation, as was originally intended with standard NVFP4 quantization with stochastic rounding.
However, we find that both options underperform the standard 4/6 formulation for quantizing gradients, providing no improvement over baseline NVFP4 training.
Our results are shown in \cref{fig:pretraining_gradients}, and we continue to use the biased 4/6 selection mechanism in the remainder of this section.

One explanation for why this may improve performance comes from the fact that even though 4/6 is biased when used with stochastic rounding, it provides the least quantization error of these three options evaluated.
It is possible that the benefits provided by the reduced quantization error outweighs those provided by the unbiased nature of stochastic rounding.
We leave this evaluation to future work.

\clearpage
\subsection{Per-Tensor Quantization}

\begin{figure}[ht]
    \centering
    \includegraphics[width=\linewidth]{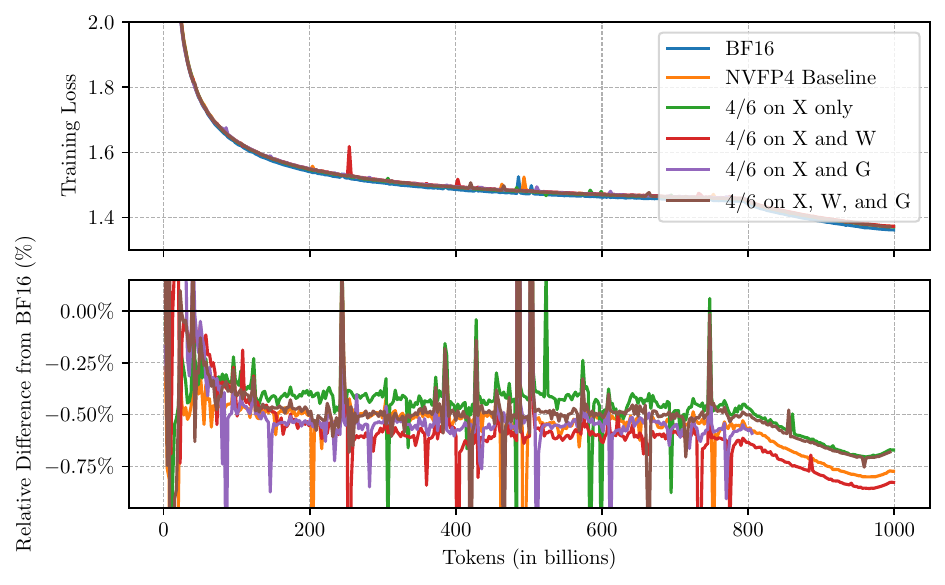}
    \caption{\textbf{Applying 4/6 to Activations Yields the Best Pre-Training Performance.} Even though benefits can be observed when all tensors are quantized using 4/6, only applying 4/6 during activation quantization provides comparable performance with less computational overhead.}
    \label{fig:pretraining_tensor_specific}
\end{figure}

Finally, we evaluate which tensors benefit the most from being quantized with 4/6.
Surprisingly, we find that when 4/6 is only applied to activations, training performance is comparable to when 4/6 is applied to all three types of tensors (activations, weights, and gradients).
These results are shown in \cref{fig:pretraining_tensor_specific}.

Notably, this finding is not replicated when 4/6 is only applied to both activations and weights, or when it is only applied to both activations and gradients.
Since 4/6 introduces a small amount of computational overhead, this finding leads to our final recommendation, which is that only activations should be quantized with 4/6 during large-scale pre-training experiments with NVFP4.

\end{document}